\documentclass[runningheads]{llncs}
\usepackage{graphicx}

\usepackage{tikz}
\usepackage{comment}
\usepackage{amsmath,amssymb, physics} 

\usepackage[accsupp]{axessibility}  

\usepackage{multirow}
\usepackage{booktabs}
\usepackage{tabularx} 

\newcolumntype{L}[1]{>{\hsize=#1\hsize\raggedright\arraybackslash}X}%
\newcolumntype{R}[1]{>{\hsize=#1\hsize\raggedleft\arraybackslash}X}%
\newcolumntype{C}[1]{>{\hsize=#1\hsize\centering\arraybackslash}X}%

\usepackage{gensymb} 
\usepackage{subfigure}
\usepackage{color,soul}

\begin{document}
\pagestyle{headings}
\mainmatter
\def\ECCVSubNumber{7838}  

\title{Gaussian Activated Neural Radiance Fields for High Fidelity Reconstruction \& Pose Estimation}

\titlerunning{GARF for High Fidelity Reconstruction \& Pose Estimation}
%
\author{Shin-Fang Chng\and
Sameera Ramasinghe\and
Jamie Sherrah\and
Simon Lucey}
\authorrunning{Shin-Fang et al.}
%
\institute{Australian Institute for Machine Learning \\
University of Adelaide\\
\email{\{shinfang.chng,sameera.ramasinghe,jamie.sherrah,simon.lucey\}@adelaide.edu.au}}
\maketitle

\begin{abstract}
Despite Neural Radiance Fields (NeRF) showing compelling results in photorealistic novel views synthesis of real-world scenes, most existing approaches require accurate prior camera poses. Although approaches for jointly recovering the radiance field and camera pose exist~\cite{lin2021barf}, they rely on a cumbersome coarse-to-fine auxiliary positional embedding to ensure good performance. We present Gaussian Activated neural Radiance Fields (GARF) a new positional embedding-free neural radiance field architecture -- employing Gaussian activations -- that outperforms the current state-of-the-art in terms of high fidelity reconstruction and pose estimation. 

\end{abstract}

\section{Introduction}


Recent work by Lin \textit{et al.} \cite{lin2021barf} -- Bundle Adjusted neural Radiance Fields (BARF) -- revealed that an architecturally-modified Neural Radiance Field (NeRF)~\cite{mildenhall2020nerf} could effectively solve the joint task of scene reconstruction and pose optimization. One crucial insight from this work is that the error backpropagation to the pose parameters in traditional NeRF is hampered by large gradients due to the high-frequency components in the positional embedding. To ameliorate this effect, the authors proposed a coarse-to-fine scheduler to gradually enable the frequency support of the positional embedding layer throughout the joint optimisation. Although achieving impressive results, this workaround requires careful tuning of the frequency scheduling process through a cumbersome \textit{multi-dimensional} parameter sweep. In this paper we investigate if this coarse-to-fine strategy can be bypassed through other means; simplifying the approach and potentially opening up new avenues for improvement. 

NeRF is probably the most popular application of coordinate multi-layer perceptrons (MLPs). NeRF maps an input 5D coordinate (3D position and 2D viewing direction) to the scene properties (view-dependent emitted radiance and volume density) of the corresponding location. A crucial ingredient of most coordinate MLPs is positional encoding. Traditional MLPs suffer from \textit{spectral-bias} -- \textit{i.e.,} they are biased towards learning low-frequency functions -- when used for signal reconstruction. Thus, MLPs, in their rudimentary form, are not ideal for encoding natural signals with fine detail, which entails modeling large fluctuations~\cite{rahimi2007random}. To circumvent this issue, NeRF architecturally modifies the MLPs by projecting the low-dimensional coordinate inputs to a higher dimensional space using a positional embedding layer, which allows NeRF to learn high-frequency components of the target function rapidly~\cite{tancik2020fourier,mildenhall2020nerf}.

Of late, there has been an increasing advocacy towards self-contained coordinate networks. Of particular note in this regard is the work of Sitzmann \textit{et al}~\cite{sitzmann2020implicit} who advocated that by simply replacing conventional activation functions (e.g. ReLU) with sine -- one can remove the need for any type of positional embedding. Although showing promise, such sine-MLPs have been found experimentally to be sensitive to weight initialization~\cite{sitzmann2020implicit,ramasinghe2021beyond}. While Sitzmann \textit{et al.}~\cite{sitzmann2020implicit} proposed an initialization scheme that aids sine-MLPs to achieve faster convergence when solving for signal reconstruction. Their deployment within NeRF has been limited, with most of the community still opting for positional embedding with conventional activations. 

\noindent \textbf{Contributions:} In this paper we draw inspiration from recent work~\cite{ramasinghe2021beyond} that has advocated for a broader class of effective activation functions -- beyond sine -- that can also circumvent the need for positional encoding. Of particular note in this regard are Gaussian activations. To our knowledge, their use in joint signal recovery and pose estimation has not been previously explored. We illustrate that these activations can preserve the first-order gradients of the target function better than conventional activations enhanced with positional embedding layers. When applied to BARF -- that is simultaneously solving for pose and radiance field reconstruction -- sine-MLPs are quite susceptible to local minima (even with good initialization), but our proposed Gaussian Activated neural Radiance Fields (GARF) exhibit robust state-of-the-art performance. 

In summary, we present the following contributions:
\begin{itemize}
    \item We present GARF, a self-contained approach for reconstructing neural radiance field from imperfect camera poses without cumbersome hyper-parameter tuning and model initialisation.
    \item We establish theoretical insights of the effect of Gaussian activation in the joint optimisation problem of neural radiance field and camera poses, supported by an extensive empirical results.
\end{itemize} 
We demonstrate that our proposed GARF can successfully recover scene representations from unknown camera poses, even on challenging scenes with low-textured region, paving the way for unlocking NeRF for real-world applications.

\section{Related Work}



\subsection{Neural Scene Representations.}
Recent works have demonstrated the potential of multi-layer perceptrons or also known as MLPs as \textit{continuous} and \textit{memory efficient} representation for 3D geometry, including shapes~\cite{genova2019learning,genova2020local}, objects~\cite{mescheder2019occupancy,chabra2020deep,park2019deepsdf} or scene~\cite{sitzmann2019scene,jiang2020local,sitzmann2020implicit}. Using 3D data such as point clouds as supervision, these approaches typically optimise signed distance functions~\cite{park2019deepsdf,jiang2020local} or binary occupancy fields~\cite{mescheder2019occupancy,genova2020local}. To alleviate the dependency of 3D training data, several methods formulate differentiable rendering functions which enables the networks to be optimised using multiview 2D images~\cite{sitzmann2019scene,niemeyer2020differentiable,mildenhall2020nerf,tewari2021advances}. Of particular interest is NeRF~\cite{mildenhall2020nerf}, which models the continuous radiance field of a scene using a coordinate-MLP in a volume rendering framework by minimising the photometric errors. Due to its simplicity and unprecedented high fidelity novel view synthesis, NeRF has attracted wide attention across the vision community~\cite{park2021nerfies,deng2021depth,martin2021nerf,turki2021mega,tancik2022block,yu2021pixelnerf}. Numerous extensions have been made on many fronts, e.g., faster training and inference~\cite{deng2021depth,yu2021plenoctrees,reiser2021kilonerf,lindell2021autoint}, deformable fields~\cite{park2021nerfies}, dynamic scene modeling~\cite{li2021neural,xian2021space,gao2021dynamic}, generalisation~\cite{wang2021ibrnet,schwarz2020graf} and pose estimation~\cite{lin2021barf,wang2021nerf,yen2021inerf,meng2021gnerf,jeong2021self,sucar2021imap,su2021nerf}.

\subsection{Positional Embedding for Pose Estimation.}
Positional embedding is an integral component of MLPs~\cite{tancik2020fourier,ramasinghe2021learning,zheng2021rethinking} which enable them to learn high frequency functions in low dimensional domain. One of the earliest roots of this approach can be traced to the work by Rahimi \textit{et al.}~\cite{rahimi2007random}, who discovered that random Fourier Features can be used to approximate an arbitrary stationary kernel function. Leveraging such an insight, Mildenhall et al~\cite{mildenhall2020nerf,tancik2020fourier} recently demonstrated that encoding input coordinates with sinusoidal allows MLPs to represent higher frequency content, enables a high-fidelity neural scene reconstruction in novel view synthesis. 

Despite the ability of positional embedding in enabling MLPs to represent high frequency components, choosing the right frequency scale which often involves a cumbersome parameter tuning is critical, \textit{i.e.}, if the bandwidth of the signal is increased excessively, coordinate-MLP tend to produce noisy signal interpolations~\cite{tancik2020fourier,ramasinghe2022regularizing,hertz2021sape}.

More recently, there has been an increasing interest in using coordinate-MLPs to tackle the joint problem of neural scene reconstruction and pose optimization~\cite{lin2021barf,wang2021nerf,yen2021inerf,meng2021gnerf,jeong2021self,sucar2021imap,su2021nerf,zhu2021nice}. Remarkably, Lin \textit{et al.}~\cite{lin2021barf} demonstrated that coordinate-MLPs entail an unanticipated drawback in camera registration -- \textit{i.e.}, large gradient due to high frequency components in positional encoding function could hamper the error backpropagation to the pose parameters. Based on this observation, they proposed a work-around to anneal each component of the frequency function in a coarse-to-fine manner. By enabling a smoother trajectory for the optimisation problem, they show that such a strategy can lead to better pose estimation, compared to \textit{full} positional encoding. Unlike BARF, we take different stance -- i.e, is there a \textit{self-contained} architecture which can tackle the pose estimation problem optimally while attaining high fidelity neural scene reconstruction without positional embedding?


\subsection{Embedding-free Coordinate-networks.}
Sitzmann \textit{et al.}~\cite{sitzmann2020implicit} alternatively proposed sinusoidal activation functions which enable coordinate MLPs to encode high frequency functions without positional embedding layer. Despite its potential, networks that employ sinusoidal activations are hyper-sensitive to the initialisation scheme~\cite{sitzmann2020implicit,ramasinghe2021beyond,ramasinghe2022regularizing}. Taking a step further, Ramasinghe \textit{et al.}, recently broadened the understanding the effect of different activations in MLPs. They proposed a class of novel \textit{non-periodic} activations that can enjoy more robust performance against random initialisation than sinuosoids. Our work significantly differs from the above-mentioned works. While we also advocate for a simple and robust embedding-free coordinate network, our work focuses on the joint problem of high fidelity neural scene reconstruction and pose estimation.




\section{Method}
In this section, we will provide an exposition of our problem formulation and different classes of coordinate networks, characterising the relative merits of each class for joint optimisation of neural scene reconstruction and pose estimation. 

\subsection{Formulation}
We first present the formulation of recovering the 3D neural radiance field from NeRF~\cite{mildenhall2020nerf} jointly with camera poses.
We denote $\mathcal{T}$ as the camera pose transformations, and $F$ as the network in NeRF, respectively. NeRF encodes the volumetric field of a 3D scene using a coordinate-network as $F:\mathbb{R}^{3} \xrightarrow{}\mathbb{R}^4$, which maps each input 3D coordinate $\mathbf{x}\in \mathbb{R}^3$ to its corresponding volume density $\sigma \in \mathbb{R}$ and directional emitted colour $\mathbf{c}\in \mathbb{R}^3$, i.e., $F(\mathbf{x}; \mathbf{\Theta}) = [\mathbf{c},\sigma]$, where $\mathbf{\Theta}$ is the network weights~\footnote{$f$ is also conditioned on viewing direction for modeling view-dependent effect, for which we omit here in the derivation for simplicity.}.

Let $\mathbf{u} \in \mathbb{R}^2$ be the pixel coordinates, $\mathcal{I}:\mathbb{R}^2 \xrightarrow{} \mathbb{R}^3$ be the imaging function. Given a set of images $\{\mathcal{I}_i\}_{i=1}^{M}$, we aim to solve for a volumetric radiance field $\mathbf{\Theta}$ of a 3D scene and the camera poses $\{\mathbf{p}_i\}_{i=1}^{M}$ by minimizing the photometric loss as
\begin{equation}\label{eq:nerf_objfn}
    \min_{\{\mathbf{p}_{i}\}_{i=1}^{M} \in \mathfrak{se}(3), \mathbf{\Theta}} \,\,{\sum_{i=1}^{M} \sum_{\mathbf{u} \in \mathbb{R}^{2}}\| \mathcal{\hat{I}}(\mathbf{u;\mathbf{p}}_{i}, \mathbf{\Theta}) - \mathcal{I}_{i}(\mathbf{u}) \|_{2}^{2}}.
\end{equation}

First, we assume the rendering operation of NeRF in the camera coordinate system. Expressing the pixel coordinate in its homogeneous coordinate as $\tilde{\mathbf{u}}$, we can define a 3D point $\mathbf{x}_i$ along a camera ray sampled at depth $t_{i}$ as $\mathbf{x}_{i} = t_{i}\tilde{\mathbf{u}}$. The estimated RGB colour of $\hat{\mathcal{I}}$ at pixel coordinate $\mathbf{u}$ is then computed by aggregating the predicted $\mathbf{c}$ and $\sigma$ as
\begin{equation}\label{eq:volume_rendering}
    \hat{\mathcal{I}}(\mathbf{u}) = \int_{t_{n}}^{t_{f}} T(\mathbf{u}, t) \sigma(t \tilde{\mathbf{u}} ) \mathbf{c} (t \tilde{\mathbf{u}}) dt 
\end{equation}
where $T(\mathbf{u}, t) = \exp(-\int_{t_{n}}^{t} \sigma(t'\tilde{\mathbf{u}} )) dt'$, and
$t_{n}$ and $t_{f}$ are the bounds of the depth range of interest; see~\cite{levoy1990efficient} for more details of volume rendering operation. In practice, the integral is commonly approximated using quadrature~\cite{mildenhall2020nerf} which evaluates the network $F$ at a discrete set of $N$ points through stratified sampling~\cite{mildenhall2020nerf} at depth $\{t_1,...,t_N\}$. Therefore, this entails $N$ querying of the network $F$, whose output $\{\mathbf{y}_{i}\}_{i=1}^N$ are composited through volume rendering. Denoting the ray compositing function as $G: \mathbb{R}^{4N} \xrightarrow{} \mathbb{R}^3$, we can rewrite $\tilde{\mathcal{I}}(\mathbf{u})$ as $\tilde{\mathcal{I}}(\mathbf{u}) = G(\mathbf{y}_1,..., \mathbf{y}_N)$. 
Given a camera pose $\mathbf{p}$, we can transform a 3D point $\mathbf{x}$ in the camera coordinate system to the world coordinate system through a 3D rigid transformation $\mathcal{T}$ to obtain the synthesized image as 
\begin{equation}
    \mathcal{\hat{I}}(\mathbf{u};\mathbf{p}) = G\bigg( \{F\big(\mathcal{T}(t_{i}\tilde{\mathbf{u}}; \mathbf{p});\mathbf{\Theta}\big)\}_{i=1}^{N}\bigg). 
\end{equation}
We solve the optimization problem~(\ref{eq:nerf_objfn}) using gradient descent. Next, we give a brief exposition of coordinate-networks and compare them.

\subsection{Coordinate-networks} 
Coordinate-networks are a special class of MLPs that are used to encode signals as trainable weights. 
An MLP with $L$ layers can be formulated as
\begin{align}
   F(\mathbf{x}) = (g^{[L]} \circ \Phi^{[L-1]} \circ g^{[L-1]} \circ \dots \Phi^{[1]} \circ g^{1})(\mathbf{x}^{1}) + \mathbf{b}^{[L]},
\end{align}
where $g^{[l]} = \mathbf{W}^{[l]}\mathbf{x}^{[l]} + \mathbf{b}^{[l]}$, $\mathbf{W}^{[l]}$ are trainable weights at the $l^{th}$ layer, $\mathbf{b}^{[l]}$ is the bias, and $\Phi^{[l]}(\cdot)$ is a non linear function. With this definition in hand, we briefly discuss several types of coordinate-networks below.

\subsubsection{ReLU-MLPs:} employ the ReLU activation function $\Phi(x) = max(0,x)$. Despite being a universal approximator in theory, ReLU-MLPs are biased towards learning low-frequency functions~\cite{xu2019training,rahaman2019spectral}, making them sub-optimal candidates for encoding natural signals with high signal fidelity. To circumvent this issue, various methods have been proposed in the literature, which we shall discuss next.
\vspace{-1mm}
\subsubsection{PE-MLPs:} are the most widely adapted class of coordinate-networks and were popularized by the seminal work of \cite{mildenhall2020nerf} through the use of positional embedding (PE). In PE-MLPs, the low-dimensional input coordinates are projected to a higher-dimensional hypersphere via a positional embedding layer $\gamma(\mathbf{x}) \in \mathbb{R}^3 \xrightarrow{} \mathbb{R}^{3+6D}$, which takes the form
\begin{equation}\label{eq:pos_embedding}
    \gamma(\mathbf{x}) = \big[\mathbf{x}, [\,\sin(2\pi\mathbf{x}), \cos(2\pi\mathbf{x})], ..., [\,\sin(2^{D-1}\pi\mathbf{x}), \cos(2^{D-1}\pi\mathbf{x})]\big],
\end{equation}
where $D$ is a hyper-parameter that controls the total number of frequency bands. After computing (\ref{eq:pos_embedding}), the embedded 3D input points are then passed through a conventional ReLU-MLPs to obtain $F(\gamma(\mathbf{x}); \mathbf \Theta)$. 
\vspace{-1mm}
\subsubsection{Sine-MLPs:} are a coordinate-network type without a positional embedding; as proposed by ~\cite{sitzmann2020implicit}. In sine-MLPs, the activation function is a sinusoid of the form
\begin{align} \label{eq:sine}
    \mathbf{x}^{[l]} \mapsto{}\Phi^{[l]}(\mathbf{x}^{[l]}) = \sin(2\pi \omega_{o} \mathbf{x}^{[l]}),
\end{align}
where $w_0$ is a hyperparameter. A larger $w_0$ increases the bandwidth of the network, allowing it to encode increasingly higher frequency functions.  
\vspace{-1mm}
\subsubsection{Gaussian-MLPs:} are a recent class of positional-embedding less coordinate-networks~\cite{ramasinghe2021beyond}, where the activation function is defined as
\begin{align*}
   \mathbf{x}^{[l]} \mapsto{}\Phi^{[l]}(\mathbf{x}^{[l]}) = \exp(\frac{-\mathbf{x}^{[l]^{2}}}{2\sigma^{2}}).
\end{align*}
Here, $\sigma$ is a hyperparameter that can be used to tune the bandwidth of the network: a larger $\sigma$ corresponds to a lower bandwidth, and vise-versa.

\subsection{GARF for Reconstruction and Pose Estimation}\label{subsec:GARF}

In this paper, we advocate the use of Gaussian-MLPs for jointly solving pose estimation and scene reconstruction, and show substantial empirical evidence that they yield better accuracy and easier optimization over the other choices. We speculate the reason for this superior performance as follows. The pose parameters are optimized using the gradients flow through the network. Hence, the ability to accurately represent the first-order derivatives of the encoded signal plays a key role in optimizing pose parameters. However, Sitzmann \textit{et al.}~\cite{sitzmann2020implicit} showed that PE-MLPs are incapable of accurately model first-order derivatives of the target signal, resulting in noisy artifacts. This impacts the Fourier spectrum of the network function, which is implicitly related to the derivatives. It was shown in  \cite{ramasinghe2022regularizing} that the Fourier transform $f(\vb{k})$ of a shallow Gaussian-MLP takes the form

\begin{equation}
\label{eq:gaussian}
  f(\vb{k}) =  \sum_{i= 1}^{m} w^{(2)}_i \frac{(2\pi)^{\frac{n+1}{2}}\sigma}{|\vb{w}^{(1)}_i|}e^{-\Big(\sqrt{2}\pi  \frac{\vb{w}^{(1)}_i}{|\vb{w}^{(1)}_i|^{2}}\cdot \vb{k} \sigma \Big)^2} \delta_{\vb{w}^{(1)}_i}(\vb{k}),
\end{equation}
where $\vb{k}$ is the frequency index,  $\delta_{\vb{w}}(\vb{k})$ is the Dirac delta distribution which concentrates along the line spanned by $\vb{w}$, and $\vb{w}^{(i)}$ are the weight vectors corresponding to the $i^{th}$ layer. Note that Eq. (\ref{eq:gaussian}) is a smooth distribution, which is parameterized by $\sigma$ and $\vb{w}^{(i)}$'s. In other words, for a suitably chosen $\sigma$, the bandwidth of the network can be increased in a continuous manner by appropriately learning the weights. Furthermore, as $\sigma$ is a continuous parameter, it provides MLPs with the ability to smoothly manipulate the spectrum of the Gaussian MLP.

In contrast, \cite{yuce2021structured} demonstrated that spectrum of a PE-MLP tends to consist of discrete spikes, where the spikes are placed on the integer harmonics of the positional embedding frequencies. Approximating the ReLU function via a polynomial in the form $\rho(x) = \sum_{i=1}^K \alpha_ix^i$, where $\alpha_i$ are constants, they showed that the spectrum is concentrated on the frequency set 
\[
 \Big\{ \sum_{d = 1}^D s_d 2^d\pi|s_d \in \mathbb{Z} \wedge \sum_{d=1}^D |s_d| < K \Big\}.
\]




Recall that in order to increase the frequency support of the positional embedding layer, one needs to increase $D$. It is evident that increasing $D$ even by one adds many harmonic spikes on the spectrum at the high-frequency end, irrespective of the network weights. Therefore, it is not possible to manipulate the spectrum of the PE-MLP continuously under a controlled setting. This can result in unnecessary high-frequency components that lead to unwanted artifacts. 

On the other hand, sine-MLPs are able to construct rich spectra and represent first-order derivatives accurately \cite{sitzmann2020implicit}. A drawback, however, is that sine-MLPs are extremely sensitive to initialization. Sitzmann \textit{et al.} \cite{sitzmann2020implicit} proposed an initialization scheme for sine-MLPs in signal reconstruction, under which they show strong convergence properties. However, we empirically demonstrate that when jointly optimizing for the pose parameters and scene reconstruction, the above initialization yields sub-par performance, making sine-MLPs highly likely to get trapped in local minima. We also show that, in comparison, Gaussian-MLPs exhibit far superior convergence properties, indicating that they entail a simpler loss landscape.

\section{Experiments}
This section validates and analyses the effectiveness of our proposed GARF with other coordinate networks. We first unfold the analysis on a 2D planar image alignment problem, and demonstrate extensive results on learning NeRF from unknown camera poses.

\begin{figure}[t]
    \centering
    {\includegraphics[height=3cm,width=10cm]{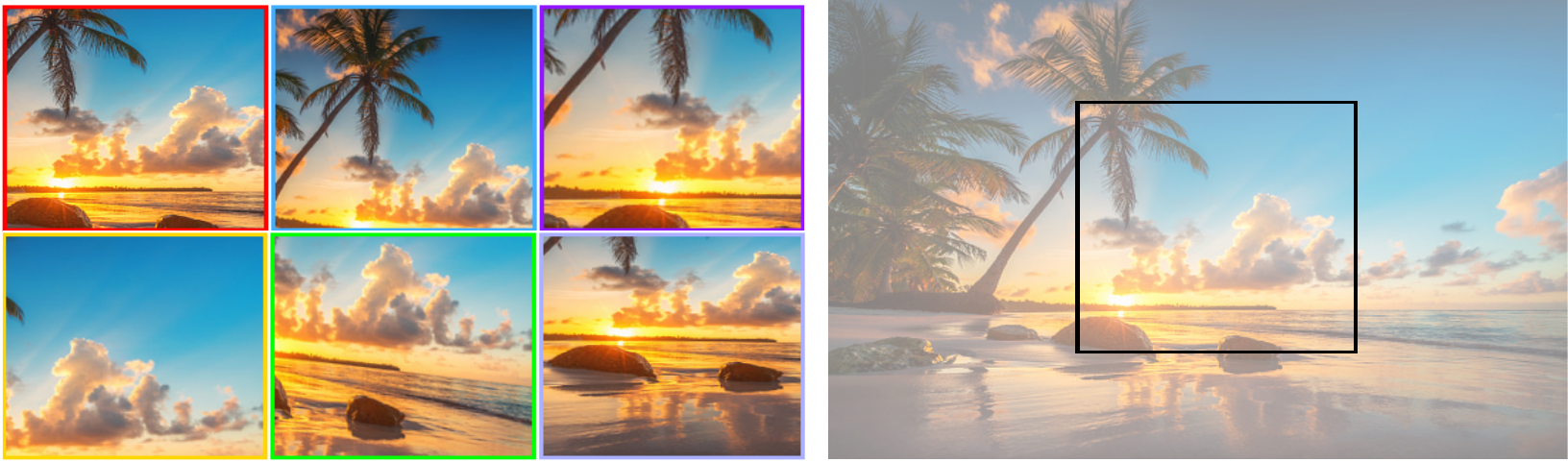}}
    \caption{A 2D planar image alignment instance. \textit{Left}: Input image patches with $N=6$. \textit{Right}: The initial poses are initialised as identity. }
    \label{fig:2d_example}
\end{figure}
\begin{figure}[t]
    \centering
    \subfigure{\includegraphics[width=\textwidth]{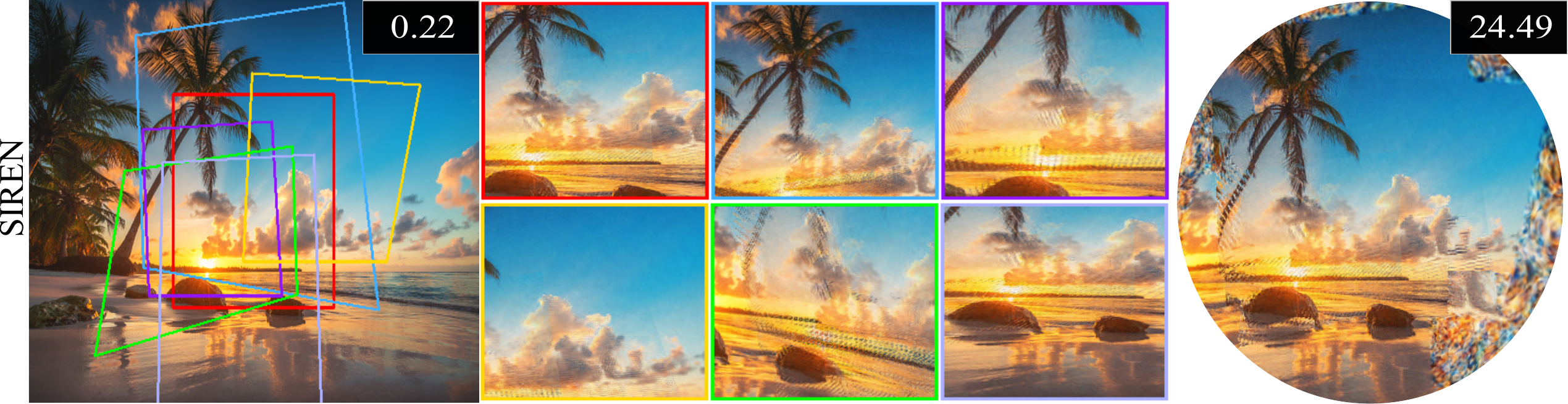}}     
    \subfigure{\includegraphics[width=\textwidth]{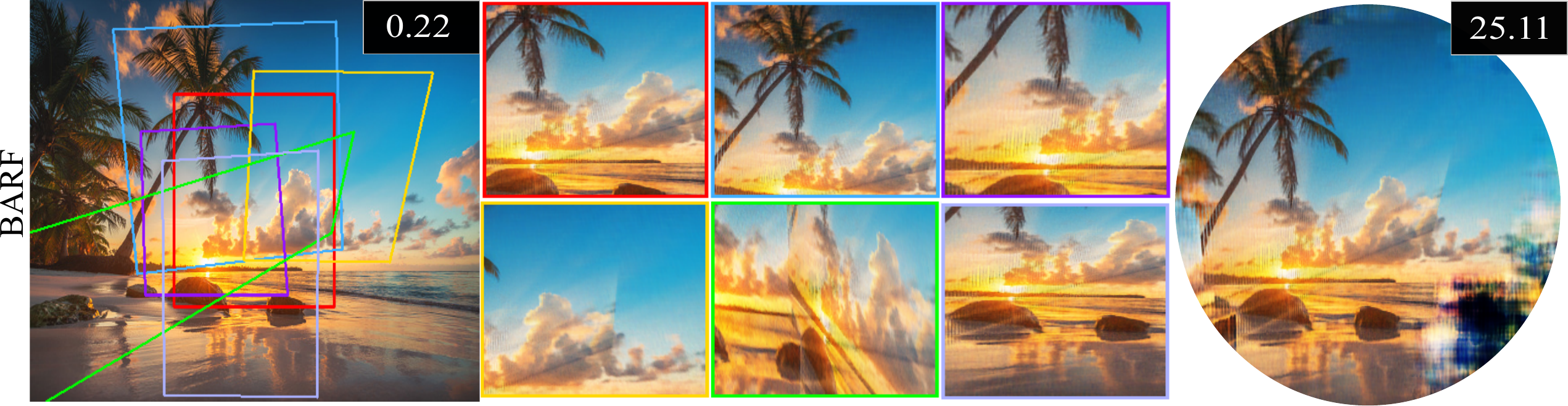}}
    \subfigure{\includegraphics[width=\textwidth]{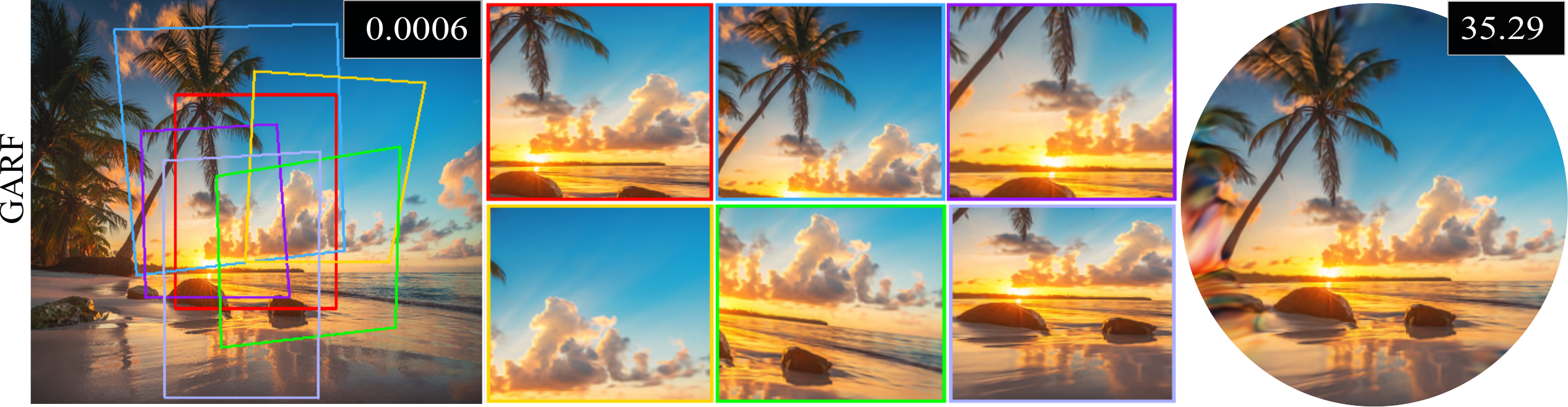}}
    \caption{Qualitative and quantitative results of the 2D planar image alignment problem. \textit{Left}: Visualisation of the estimated poses with the $\mathfrak{sl}(3)$ error. \textit{Center}: Reconstruction of each warped patches.  \textit{Right}: Final image reconstruction with the patch PSNR.}
    \label{fig:2d_sunset}
\end{figure}


\begin{figure}[t]
    \centering
    \includegraphics[width=\textwidth]{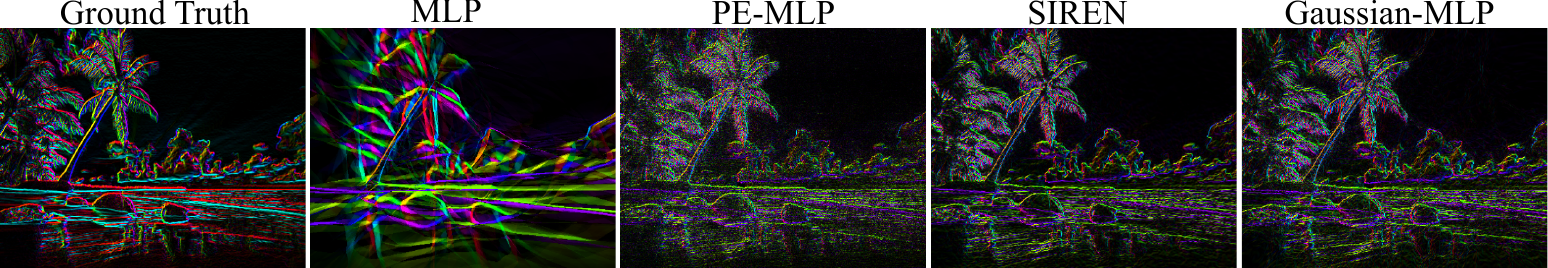}
    \caption{Comparison of the first-order derivatives of encoded signal $\nabla{F}$ on solving an image reconstruction problem. The first-order derivative of each function is computed using network's output with respect to the coordinates. Note that \textit{only} the groundtruth derivative is computed using Sobel Filter. }
    \label{fig:2d_sunset_derivatives_img} 
    \end{figure}

\begin{figure}[t]
    \centering
    \includegraphics[width=\textwidth]{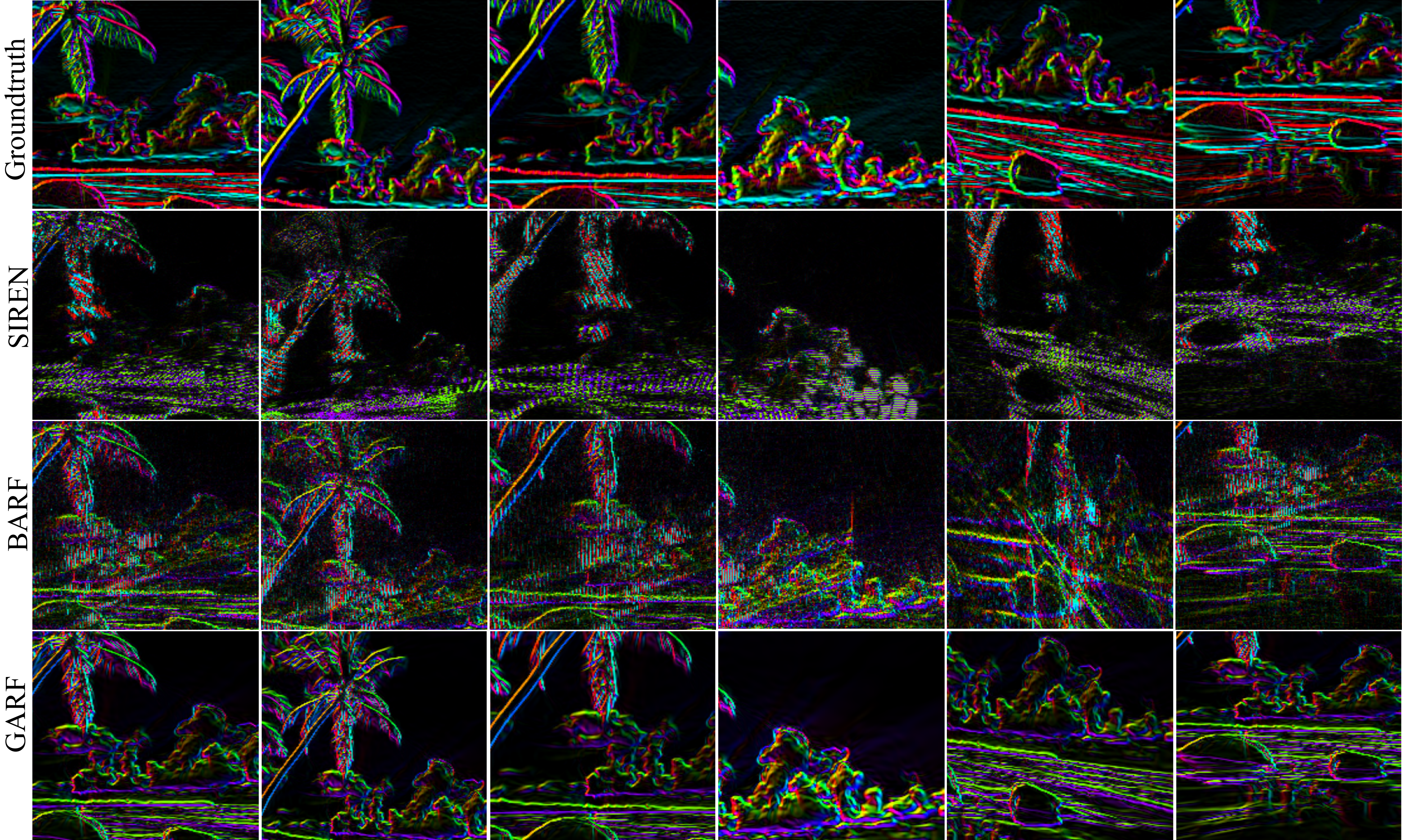}
    \caption{Comparison of the first-order derivatives of encoded signal $\nabla{F}$. The first-order derivative of each function is computed using network's output with respect to the coordinates. Note that \textit{only} the groundtruth derivative is computed using Sobel Filter. }
    \label{fig:2d_sunset_derivatives} 
    \end{figure}
    
\begin{figure}
    \centering
    {\includegraphics[width=\textwidth]{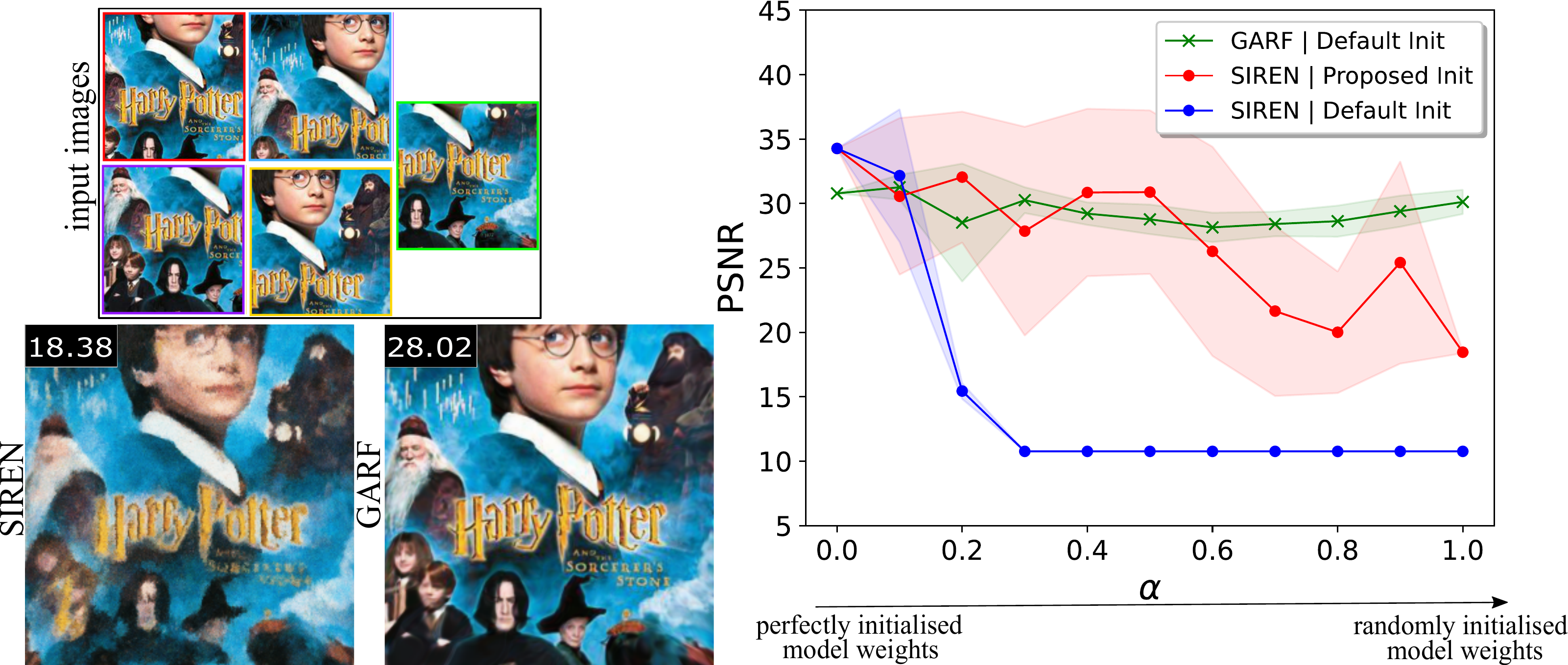}}
    \caption{\textit{Left}: Input images and the image reconstruction for SIREN and GARF, which correspond to the \textit{red} and \textit{green} curve, respectively. \textit{Right}: Robustness of the initialisation at different $\alpha$. When $\alpha=0$, all the networks are initialised with \textit{optimal} weights; When $\alpha=1$, all the networks are initialised with \textit{random} weights. Note that for SIREN, we also investigate the case when SIREN strictly adheres to the initialisation scheme proposed by Sitzmann \textit{et al.}~\cite{sitzmann2020implicit} (\textit{red}). The shaded areas correspond to the two standard deviations over 10 runs.}
    \label{fig:siren_vs_gaussian_init}
\end{figure}

\subsection{2D Planar Image Alignment}
 To develop intuition, we first consider the case of 2D planar image alignment problem. More specifically, let $\mathbf{u} \in \mathbb{R}^2$ be the 2D pixel coordinates and $\mathcal{I}:\mathbb{R}^2 \rightarrow \mathbb{R}^3$, we aim to optimize a neural image representation parameterised as the weights of coordinate network $F$ while also solving for warp parameters as
 \begin{align}\label{eq:2d_alignment}
    \min_{\{\mathbf{p}_{i}\}_{i=1}^{N} \in \mathfrak{sl}(3), \mathbf{\Theta}} \,\,{\sum_{i=1}^{N} \sum_{\mathbf{u} \in \mathbb{R}^{2}}\| F(\mathcal{W}(\mathbf{u;\mathbf{p}}_{i});\mathbf{\Theta}) - \mathcal{I}_{i}(\mathbf{u}) \|_{2}^{2}},
\end{align}
where $\mathcal{W}:\mathbb{R}^2 \xrightarrow{} \mathbb{R}^{2}$ denotes the warp function parameterised as $\mathbf{p} \in \mathfrak{sl}(3)$.
Given $N=6$ patches from the image $\mathcal{I}$ generated with random homography perturbations, we aim to jointly estimate the \textit{unknown} homography warp parameters ${\mathbf{p}_i}$ and network weights $\mathbf{\Theta}$. We fix the gauge freedom by anchoring the first patch as identity; see Fig.~\ref{fig:2d_example} for an example. 

\subsubsection{Experimental settings.} 
We compare our proposed GARF with the following networks: PE-MLP with a coarse-to-fine embedding annealer (BARF)~\cite{lin2021barf} and sine-MLP (SIREN)~\cite{sitzmann2020implicit}. We use a $4$-layer MLP with 256-dimensional hidden units for all networks. We use the Adam optimizer to optimize both the network weights ${\mathbf{\Theta}}$ and the warp parameters $\mathbf{p}$. We use a learning rate that begins at $1\times 10^{-3}$ for ${\mathbf{\Theta}}$, and $3\times10^{-3}$ for $\mathbf{p}$, with both decaying exponentially to $1 \times 10^{-4}$ for GARF and BARF. For SIREN, we use a learning rate of $1 \times 10^{-4}$ for both ${\mathbf{\Theta}}$ and $\mathbf{p}$ decaying exponentially to $1\times10^{-5}$. For BARF, we use $D=8$ frequency bands (\ref{eq:pos_embedding}), and linearly anneal the frequency from $0$ to $8$ over $2$K iterations. Note that we use the same parameters as proposed in \cite{lin2021barf}. At each optimization step, we randomly sample $15\%$ of the pixel coordinates for each patch.

\subsubsection{Initialisation.}\label{subsubsec:init}
For BARF and SIREN, we use the initialisation scheme proposed in the original paper~\cite{mildenhall2020nerf,lin2021barf,sitzmann2020implicit}, whereas for our proposed GARF we simply use randomly initialised weights. We initialise the warp parameters $\{\mathbf{p}_{i}\}_{i=1}^{N}$ as identity for all models; see Fig.~\ref{fig:2d_example}.

\subsubsection{Results.}
We show the quantitative and qualitative registration results in Fig.~\ref{fig:2d_sunset}. As GARF is able to correctly estimate the warp parameters of all patches, GARF can reconstruct the image with high fidelity. On the other hand, BARF and SIREN struggle with the image reconstruction due to misalignment.
It is important to note that the Gaussian-MLP initialisation protocol put the proposed method at a disadvantage. This further demonstrates the robustness of Gaussian-MLP towards initialisation. 

\subsubsection{First-order derivatives analysis.}
For completeness, we first inspect the first-order derivations of each coordinate network when solving for an image reconstruction task as $ \min_{\mathbf{\Theta}} \,\,{\sum_{\mathbf{u} \in \mathbb{R}^{2}}\| F(\mathbf{u};\mathbf{\Theta}) - \mathcal{I}(\mathbf{u}) \|_{2}^{2}}$; note that we use the same notations as in Eq.(\ref{eq:2d_alignment}).
As discussed in Sec.~\ref{subsec:GARF}, the ability to accurately represent the first-order derivatives of the encoded signal plays a crucial role in optimizing pose parameters. Fig.~\ref{fig:2d_sunset_derivatives_img} reinforces that the first-order derivative of the encoded signal of PE-MLP has a lot of noise artifacts -- results in poor error backpropagation to pose parameters.
While properly-initialised SIREN is capable of representing the derivatives of the signal when solving for signal reconstruction, the initialisation strategy of sine-activation is sub-optimal when jointly optimizing for neural image reconstruction and warp. As a result, the resulting function derivative is no longer well-defined; see Fig.~\ref{fig:2d_sunset_derivatives}. In contrast, GARF exhibit far superior convergence properties, albeit the model weights are initialised randomly. 


\subsubsection{Robustness of initialisation scheme.}
Additionally, we run a simple experiment to investigate the sensitivity of SIREN and GARF to initialisation. We denote ${\mathbf{\Theta}}^{*}$ as the optimal model weights, which is obtained by solving Eq.~(\ref{eq:2d_alignment}) for a neural image representation by fixing the warp parameters, and ${\bar{\mathbf{\Theta}}}$ as the randomly initialised model weights, \textit{i.e.}, weights are initialised using PyTorch default initialisation. Our goal is to solve the joint optimisation problem Eq.~(\ref{eq:2d_alignment}) by initialising $\mathbf{\Theta}$ with different scaled model weights, \textit{i.e.,} $\alpha{\bar{\mathbf{\Theta}}}+(1-\alpha){\mathbf{\Theta}}^{*}$ by linearly adjusting $\alpha$. As shown in Fig~\ref{fig:siren_vs_gaussian_init}, GARF (\textit{green curve}) is marginally affected by the initialisation, while SIREN (\textit{blue curve}) fails drastically (starting from $\alpha$=0.3). When SIREN is initialised carefully using the initialisation scheme proposed by Sitzmann \textit{et al.}~\cite{sitzmann2020implicit} (\textit{red curve}), its performance decreases as $\alpha$ gradually increases, \textit{i.e.}, as the perturbation to the optimal model weights increases. Note that the variance of performance in the GARF is much smaller compared to SIREN. 

\subsubsection{Generalisation of coarse-to-fine scheduling}
We exhaustively search through the log-space for the optimal coarse-to-fine schedulers for BARF; see supp. material for more details. The optimal coarse-to-fine hyper-parameters for each image are data-dependent, \textit{i.e.}, the hyper-paramaters tuned for one image may not be optimal for another image. In contrast to multi-dimensional coarse-to-fine schedulers, Gaussian activation function involves only one-dimensional search space, \textit{i.e.}, $\sigma$ in (\ref{eq:gaussian}).




\subsection{3D NeRF: Real World Scenes}
This section investigates the task of jointly learning neural 3D representations with NeRF~\cite{mildenhall2020nerf} on real world scenes where the camera poses are \textit{unknown}. We evaluate all the methods on the standard benchmark LLFF dataset~\cite{mildenhall2019local}, which consists of 8 real world forward-facing scenes captured by hand-held cameras.

\subsubsection{Experimental Settings.}
We compare our proposed GARF with BARF and reference NeRF (ref-NeRF). As we empirically observe that PE-MLP with scheduler (BARF) achieves better performance compared to PE-MLP~\cite{wang2021nerf} in the joint optimisation of neural radiance field and camera poses, we opted not to include the comparisons with PE-MLP here; see ~\cite{lin2021barf} or supp. for comparisons with PE-MLP. We parameterise the camera poses with the $\mathfrak{se}(3)$ Lie algebra and initialise them as \textit{identity} for GARF and BARF. We assume known intrinsics.

\subsection{Implementation Details.}
We implement our framework following the settings from \cite{mildenhall2020nerf,lin2021barf} with some modifications. For simplicity, we train a single 6-layer MLP with 256 hidden units in each layer and \textit{without hierarchical sampling}. We resize the images to $480 \times 640$ pixels and randomly sample 2048 pixel rays every iteration, each sampled at $N=128$ coordinates. We use the Adam optimizer~\cite{kingma2014adam} and train all models for 200K iterations, with a learning rate that begins at $1\times10^{-4}$ decaying exponentially to $5\times10^{-5}$, and $3\times10^{-3}$ for the poses $\mathbf{p}$ decaying to $1\times10^{-5}$. We use the default coarse-to-fine scheduling for BARF~\cite{lin2021barf}.  We use the same network size and sampling strategy for all the methods throughout our evaluation. Note that for BARF and ref-NeRF, we use the implementation from BARF; all the hyperparameters are configured as per proposed in the paper.

\subsubsection{Evaluation Details.}
We evaluate the performance of each method in terms of pose accuracy for registration and view synthesis quality for the scene reconstruction. Following \cite{lin2021barf,wang2021nerf}, we evaluate the pose error by aligning optimized poses to groundtruth via Proscustes analysis which computes the similarity transformation Sim(3) between them. Note that as the ``groundtruth'' camera poses provided in LLFF real-world scenes are the estimations from Colmap~\cite{schonberger2016structure}, the pose accuracy is only an indicator how well the estimations agree with the classical method. We report the mean rotation and translation errors for pose, as well as PSNR, SSIM and LPIPS~\cite{mildenhall2020nerf} for view synthesis in Table~\ref{table:benchmark-llff}.

\begin{figure}
    \centering
    \includegraphics[width=\textwidth]{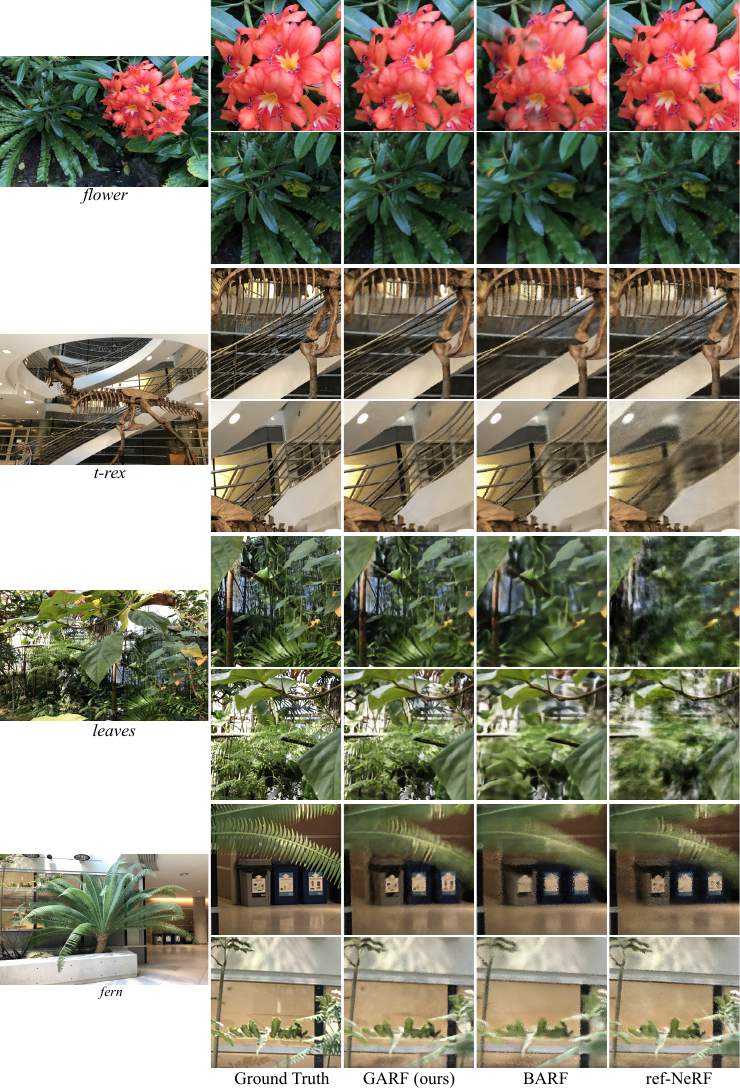}
    \caption{Qualitative results on test-views of real world scenes~\cite{mildenhall2019local}. While BARF and GARF can jointly optimize pose and the scene representation, GARF produces results with higher fidelity. Note that we use 6-layer MLPs for all methods in this experiment.}
    \label{fig:llff_qualitative}
\end{figure}

\setlength{\tabcolsep}{0.78mm}
\begin{table}[t]
    \centering
    \caption{Quantitative comparison of GARF (Ours), BARF~\cite{lin2021barf} and ref-NeRF on real-world scenes~\cite{mildenhall2019local} given \textit{unknown} camera poses.}
    \label{table:benchmark-llff}
        \begin{tabular}{l|cc|cc|cc|c|cc|c|cc|c}
            \hline\noalign{\smallskip}
            \multicolumn{1}{c|}{\textbf{Scene}} &
            \multicolumn{4}{c|}{\textbf{Pose accuracy}} &
            \multicolumn{9}{c}{\textbf{View synthesis}} \\
            \hline
            \multicolumn{1}{c|}{} &
            \multicolumn{2}{c|}{Rotation} &
            \multicolumn{2}{c|}{Translation} &
            \multicolumn{3}{c|}{PSNR $\uparrow$} &
            \multicolumn{3}{c|}{SSIM $\uparrow$} &
            \multicolumn{3}{c}{LPIPS $\downarrow$}  \\

            \multicolumn{1}{c|}{} &
            \multicolumn{2}{c|}{($\degree$)} &
            \multicolumn{2}{c|}{($10^{^{-2}}$)} &
            \multicolumn{3}{c|}{} &
            \multicolumn{3}{c|}{} &
            \multicolumn{3}{c}{}  \\
            \multicolumn{1}{c|}{} & 
            \multicolumn{1}{c}{} & \multicolumn{1}{c|}{}  &
            \multicolumn{1}{c}{} & \multicolumn{1}{c|}{}  &
            \multicolumn{1}{c}{} & \multicolumn{1}{c|}{} & \multicolumn{1}{c|}{ref-} &
            \multicolumn{1}{c}{} & \multicolumn{1}{c|}{} & \multicolumn{1}{c|}{ref-} &
            \multicolumn{1}{c}{} & \multicolumn{1}{c|}{} & \multicolumn{1}{c}{ref-} \\            
            
            \multicolumn{1}{c|}{} & 
            \multicolumn{1}{c}{\cite{lin2021barf}} & \multicolumn{1}{c|}{Ours}  &
            \multicolumn{1}{c}{\cite{lin2021barf}} & \multicolumn{1}{c|}{Ours}  &
            \multicolumn{1}{c}{\cite{lin2021barf}} & \multicolumn{1}{c|}{Ours} & \multicolumn{1}{c|}{NeRF} &
            \multicolumn{1}{c}{\cite{lin2021barf}} & \multicolumn{1}{c|}{Ours} & \multicolumn{1}{c|}{NeRF} &
            \multicolumn{1}{c}{\cite{lin2021barf}} & \multicolumn{1}{c|}{Ours} & \multicolumn{1}{c}{NeRF} \\
            \hline
            \textit{flower} & 0.47 & \textbf{0.46} & 0.25 & \textbf{0.22}
            & 23.58 & \textbf{26.40} & 23.20
            & 0.67 & \textbf{0.79} & 0.66
            & 0.27 & \textbf{0.11} & 0.27\\
            \textit{fern} & \textbf{0.16} & 0.47 & \textbf{0.20} & 0.25
            & 23.53 & \textbf{24.51} & 23.10 
            & 0.69 & \textbf{0.74} & 0.71
            & 0.34 & \textbf{0.29} & \textbf{0.29} \\
            \textit{leaves} & 1.00 & \textbf{0.13} & 0.30 & \textbf{0.23} 
            & 18.15 & \textbf{19.72} & 14.42
            & 0.48 & \textbf{0.61} & 0.24
            & 0.40 & \textbf{0.27} & 0.58 \\
            \textit{horns} & 3.50 & \textbf{0.03} & 1.32 & \textbf{0.21}
            & 19.73 & \textbf{22.54} & 19.93
            & 0.66 & \textbf{0.69} & 0.59 
            & 0.35 & \textbf{0.33} & 0.45 \\
            \textit{trex} & \textbf{0.42} & 0.66 & \textbf{0.36} & 0.48 
            & 22.63  & \textbf{22.86} & 21.42
            & 0.75 & \textbf{0.80} & 0.69
            & 0.24 & \textbf{0.19} & 0.32 \\
            \textit{orchids} & 0.71 & \textbf{0.43} & 0.42 & \textbf{0.41}
            & 19.14 &  \textbf{19.37} & 16.54
            & 0.55 & \textbf{0.57} & 0.46
            & 0.33 & \textbf{0.26} & 0.37 \\ 
            \textit{fortress} & 0.17 & \textbf{0.03} & 0.32 & \textbf{0.27} 
            & 28.48 & \textbf{29.09} & 25.62
            & 0.80 & \textbf{0.82} & 0.78
            & 0.16 & \textbf{0.15} & 0.19 \\
            \textit{room} & \textbf{0.27} & 0.42 & \textbf{0.20} & 0.32 
            & 31.43 & \textbf{31.90} & 31.65
            & 0.93 & \textbf{0.94} & 0.94
            & 0.11 & 0.13  & \textbf{0.09} \\
\hline
    \end{tabular}
\end{table}

\subsubsection{Results.}
Table~\ref{table:benchmark-llff} quantitatively contrasts the performance of GARF, BARF and ref-NeRF. As evident, Gaussian activations enable GARF to recover camera poses which matches the camera poses from off-the-shelf SfM methods. Moreover, even with shallower network, Gaussian activations can successfully recover the 3D scene representation with higher fidelity in the absence of positional embedding, compared to BARF and ref-NeRF; see the qualitative results in Fig.~\ref{fig:llff_qualitative}. 

\subsection{Real-World Demo}
\begin{figure}
    \centering
    \includegraphics[width=\textwidth]{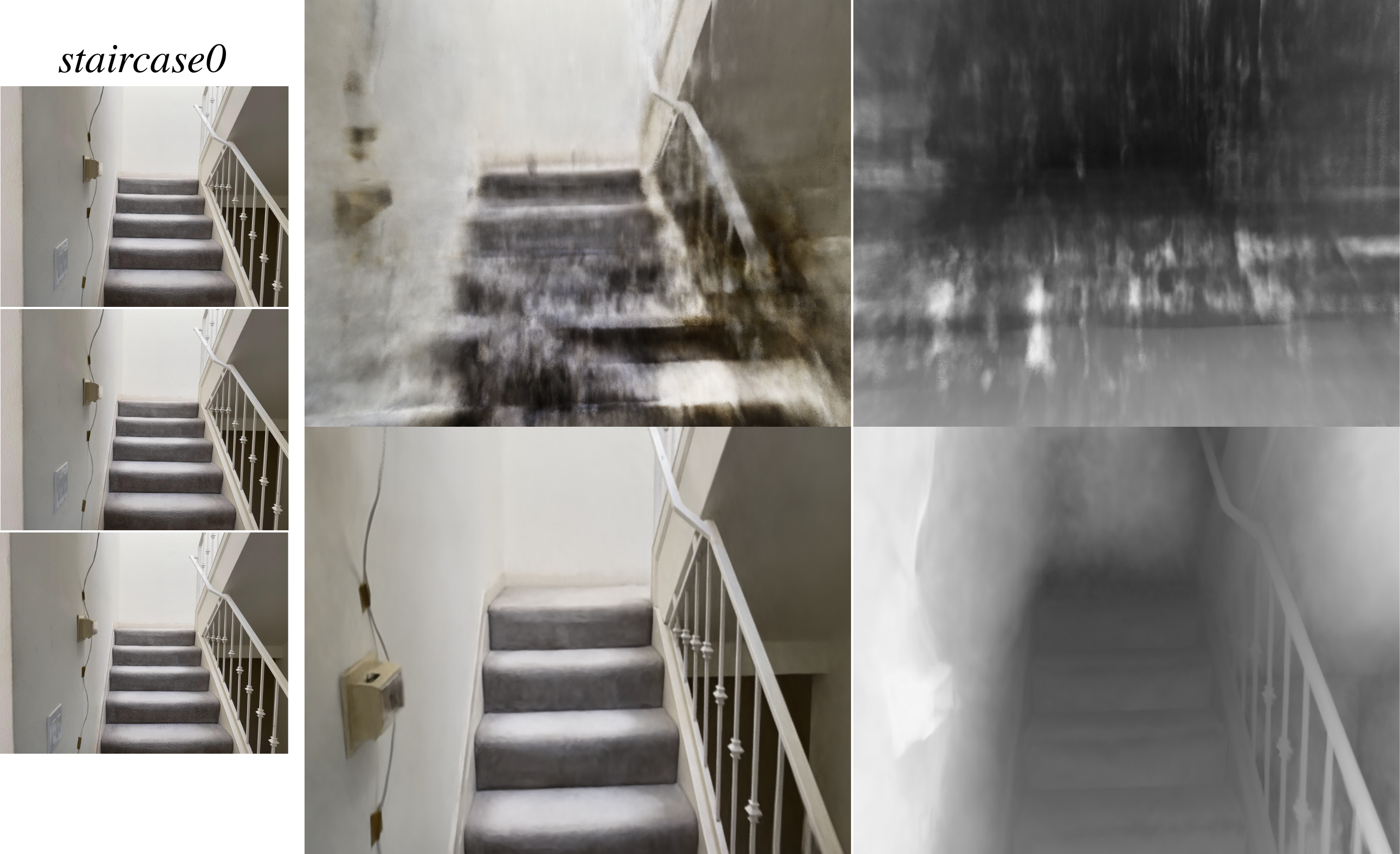}
    \caption{Novel view synthesis result on a low-textured scene captured using iPhone. \textit{Left banner}: Training images. \textit{Top row}: Rendered image and depth using ref-NeRF. \textit{Bottom row}: Rendered image and depth using GARF.}
    \label{fig:iphone_nerf}
\end{figure}

To showcase the practicability of GARF, we take one step further to test it on images of low-textured scene captured using an iPhone. Fig.~\ref{fig:iphone_nerf} remarkably demonstrate the potential of GARF on a scene with a lot of low-textured region while ref-NeRF exhibits artifacts on the novel view due to existence of outliers in front-end of SfM pipeline, which results in unreliable camera pose estimations; see supp. for more results.

\section{Conclusions}
We present Gaussian Activated neural Radiance Fields (GARF), a new positional embedding-free neural radiance field architecture that can reconstruct high fidelity neural radiance fields from imperfect camera poses without cumbersome hyper-parameter and model initialisation. 
By establishing theoretical intuition, we demonstrate that the ability of the model to preserve the first-order gradients of the target function plays an imperative role in the joint problem of optimizing for pose and radiance field reconstruction. 
Experimental results reinforced our theoretical intuition and demonstrated the superiority of GARF, even on challenging scenes with low textured region.

\clearpage
%
%
\bibliographystyle{splncs04}
\bibliography{egbib}

\begin{thebibliography}{10}
\providecommand{\url}[1]{\texttt{#1}}
\providecommand{\urlprefix}{URL }
\providecommand{\doi}[1]{https://doi.org/#1}

\bibitem{chabra2020deep}
Chabra, R., Lenssen, J.E., Ilg, E., Schmidt, T., Straub, J., Lovegrove, S.,
  Newcombe, R.: Deep local shapes: Learning local sdf priors for detailed 3d
  reconstruction. In: European Conference on Computer Vision. pp. 608--625.
  Springer (2020)

\bibitem{deng2021depth}
Deng, K., Liu, A., Zhu, J.Y., Ramanan, D.: Depth-supervised nerf: Fewer views
  and faster training for free. arXiv preprint arXiv:2107.02791  (2021)

\bibitem{gao2021dynamic}
Gao, C., Saraf, A., Kopf, J., Huang, J.B.: Dynamic view synthesis from dynamic
  monocular video. In: Proceedings of the IEEE/CVF International Conference on
  Computer Vision. pp. 5712--5721 (2021)

\bibitem{genova2020local}
Genova, K., Cole, F., Sud, A., Sarna, A., Funkhouser, T.: Local deep implicit
  functions for 3d shape. In: Proceedings of the IEEE/CVF Conference on
  Computer Vision and Pattern Recognition. pp. 4857--4866 (2020)

\bibitem{genova2019learning}
Genova, K., Cole, F., Vlasic, D., Sarna, A., Freeman, W.T., Funkhouser, T.:
  Learning shape templates with structured implicit functions. In: Proceedings
  of the IEEE/CVF International Conference on Computer Vision. pp. 7154--7164
  (2019)

\bibitem{hertz2021sape}
Hertz, A., Perel, O., Giryes, R., Sorkine-Hornung, O., Cohen-Or, D.: Sape:
  Spatially-adaptive progressive encoding for neural optimization. Advances in
  Neural Information Processing Systems  \textbf{34} (2021)

\bibitem{jeong2021self}
Jeong, Y., Ahn, S., Choy, C., Anandkumar, A., Cho, M., Park, J.:
  Self-calibrating neural radiance fields. In: Proceedings of the IEEE/CVF
  International Conference on Computer Vision. pp. 5846--5854 (2021)

\bibitem{jiang2020local}
Jiang, C., Sud, A., Makadia, A., Huang, J., Nie{\ss}ner, M., Funkhouser, T.,
  et~al.: Local implicit grid representations for 3d scenes. In: Proceedings of
  the IEEE/CVF Conference on Computer Vision and Pattern Recognition. pp.
  6001--6010 (2020)

\bibitem{kingma2014adam}
Kingma, D.P., Ba, J.: Adam: A method for stochastic optimization. arXiv
  preprint arXiv:1412.6980  (2014)

\bibitem{levoy1990efficient}
Levoy, M.: Efficient ray tracing of volume data. ACM Transactions on Graphics
  (TOG)  \textbf{9}(3),  245--261 (1990)

\bibitem{li2021neural}
Li, Z., Niklaus, S., Snavely, N., Wang, O.: Neural scene flow fields for
  space-time view synthesis of dynamic scenes. In: Proceedings of the IEEE/CVF
  Conference on Computer Vision and Pattern Recognition. pp. 6498--6508 (2021)

\bibitem{lin2021barf}
Lin, C.H., Ma, W.C., Torralba, A., Lucey, S.: Barf: Bundle-adjusting neural
  radiance fields. In: Proceedings of the IEEE/CVF International Conference on
  Computer Vision. pp. 5741--5751 (2021)

\bibitem{lindell2021autoint}
Lindell, D.B., Martel, J.N., Wetzstein, G.: Autoint: Automatic integration for
  fast neural volume rendering. In: Proceedings of the IEEE/CVF Conference on
  Computer Vision and Pattern Recognition. pp. 14556--14565 (2021)

\bibitem{martin2021nerf}
Martin-Brualla, R., Radwan, N., Sajjadi, M.S., Barron, J.T., Dosovitskiy, A.,
  Duckworth, D.: Nerf in the wild: Neural radiance fields for unconstrained
  photo collections. In: Proceedings of the IEEE/CVF Conference on Computer
  Vision and Pattern Recognition. pp. 7210--7219 (2021)

\bibitem{meng2021gnerf}
Meng, Q., Chen, A., Luo, H., Wu, M., Su, H., Xu, L., He, X., Yu, J.: Gnerf:
  Gan-based neural radiance field without posed camera. In: Proceedings of the
  IEEE/CVF International Conference on Computer Vision. pp. 6351--6361 (2021)

\bibitem{mescheder2019occupancy}
Mescheder, L., Oechsle, M., Niemeyer, M., Nowozin, S., Geiger, A.: Occupancy
  networks: Learning 3d reconstruction in function space. In: Proceedings of
  the IEEE/CVF Conference on Computer Vision and Pattern Recognition. pp.
  4460--4470 (2019)

\bibitem{mildenhall2019local}
Mildenhall, B., Srinivasan, P.P., Ortiz-Cayon, R., Kalantari, N.K.,
  Ramamoorthi, R., Ng, R., Kar, A.: Local light field fusion: Practical view
  synthesis with prescriptive sampling guidelines. ACM Transactions on Graphics
  (TOG)  \textbf{38}(4),  1--14 (2019)

\bibitem{mildenhall2020nerf}
Mildenhall, B., Srinivasan, P.P., Tancik, M., Barron, J.T., Ramamoorthi, R.,
  Ng, R.: Nerf: Representing scenes as neural radiance fields for view
  synthesis. In: European conference on computer vision. pp. 405--421. Springer
  (2020)

\bibitem{niemeyer2020differentiable}
Niemeyer, M., Mescheder, L., Oechsle, M., Geiger, A.: Differentiable volumetric
  rendering: Learning implicit 3d representations without 3d supervision. In:
  Proceedings of the IEEE/CVF Conference on Computer Vision and Pattern
  Recognition. pp. 3504--3515 (2020)

\bibitem{park2019deepsdf}
Park, J.J., Florence, P., Straub, J., Newcombe, R., Lovegrove, S.: Deepsdf:
  Learning continuous signed distance functions for shape representation. In:
  Proceedings of the IEEE/CVF Conference on Computer Vision and Pattern
  Recognition. pp. 165--174 (2019)

\bibitem{park2021nerfies}
Park, K., Sinha, U., Barron, J.T., Bouaziz, S., Goldman, D.B., Seitz, S.M.,
  Martin-Brualla, R.: Nerfies: Deformable neural radiance fields. In:
  Proceedings of the IEEE/CVF International Conference on Computer Vision. pp.
  5865--5874 (2021)

\bibitem{rahaman2019spectral}
Rahaman, N., Baratin, A., Arpit, D., Draxler, F., Lin, M., Hamprecht, F.,
  Bengio, Y., Courville, A.: On the spectral bias of neural networks. In:
  International Conference on Machine Learning. pp. 5301--5310. PMLR (2019)

\bibitem{rahimi2007random}
Rahimi, A., Recht, B.: Random features for large-scale kernel machines.
  Advances in neural information processing systems  \textbf{20} (2007)

\bibitem{ramasinghe2021beyond}
Ramasinghe, S., Lucey, S.: Beyond periodicity: Towards a unifying framework for
  activations in coordinate-mlps. arXiv preprint arXiv:2111.15135  (2021)

\bibitem{ramasinghe2021learning}
Ramasinghe, S., Lucey, S.: Learning positional embeddings for coordinate-mlps.
  arXiv preprint arXiv:2112.11577  (2021)

\bibitem{ramasinghe2022regularizing}
Ramasinghe, S., MacDonald, L., Lucey, S.: On regularizing coordinate-mlps.
  arXiv preprint arXiv:2202.00790  (2022)

\bibitem{reiser2021kilonerf}
Reiser, C., Peng, S., Liao, Y., Geiger, A.: Kilonerf: Speeding up neural
  radiance fields with thousands of tiny mlps. In: Proceedings of the IEEE/CVF
  International Conference on Computer Vision. pp. 14335--14345 (2021)

\bibitem{schonberger2016structure}
Schonberger, J.L., Frahm, J.M.: Structure-from-motion revisited. In:
  Proceedings of the IEEE conference on computer vision and pattern
  recognition. pp. 4104--4113 (2016)

\bibitem{schwarz2020graf}
Schwarz, K., Liao, Y., Niemeyer, M., Geiger, A.: Graf: Generative radiance
  fields for 3d-aware image synthesis. Advances in Neural Information
  Processing Systems  \textbf{33},  20154--20166 (2020)

\bibitem{sitzmann2020implicit}
Sitzmann, V., Martel, J., Bergman, A., Lindell, D., Wetzstein, G.: Implicit
  neural representations with periodic activation functions. Advances in Neural
  Information Processing Systems  \textbf{33},  7462--7473 (2020)

\bibitem{sitzmann2019scene}
Sitzmann, V., Zollh{\"o}fer, M., Wetzstein, G.: Scene representation networks:
  Continuous 3d-structure-aware neural scene representations. Advances in
  Neural Information Processing Systems  \textbf{32} (2019)

\bibitem{su2021nerf}
Su, S.Y., Yu, F., Zollhoefer, M., Rhodin, H.: A-nerf: Surface-free human 3d
  pose refinement via neural rendering. arXiv preprint arXiv:2102.06199  (2021)

\bibitem{sucar2021imap}
Sucar, E., Liu, S., Ortiz, J., Davison, A.J.: imap: Implicit mapping and
  positioning in real-time. In: Proceedings of the IEEE/CVF International
  Conference on Computer Vision. pp. 6229--6238 (2021)

\bibitem{tancik2022block}
Tancik, M., Casser, V., Yan, X., Pradhan, S., Mildenhall, B., Srinivasan, P.P.,
  Barron, J.T., Kretzschmar, H.: Block-nerf: Scalable large scene neural view
  synthesis. arXiv preprint arXiv:2202.05263  (2022)

\bibitem{tancik2020fourier}
Tancik, M., Srinivasan, P., Mildenhall, B., Fridovich-Keil, S., Raghavan, N.,
  Singhal, U., Ramamoorthi, R., Barron, J., Ng, R.: Fourier features let
  networks learn high frequency functions in low dimensional domains. Advances
  in Neural Information Processing Systems  \textbf{33},  7537--7547 (2020)

\bibitem{tewari2021advances}
Tewari, A., Thies, J., Mildenhall, B., Srinivasan, P., Tretschk, E., Wang, Y.,
  Lassner, C., Sitzmann, V., Martin-Brualla, R., Lombardi, S., et~al.: Advances
  in neural rendering. arXiv preprint arXiv:2111.05849  (2021)

\bibitem{turki2021mega}
Turki, H., Ramanan, D., Satyanarayanan, M.: Mega-nerf: Scalable construction of
  large-scale nerfs for virtual fly-throughs. arXiv preprint arXiv:2112.10703
  (2021)

\bibitem{wang2021ibrnet}
Wang, Q., Wang, Z., Genova, K., Srinivasan, P.P., Zhou, H., Barron, J.T.,
  Martin-Brualla, R., Snavely, N., Funkhouser, T.: Ibrnet: Learning multi-view
  image-based rendering. In: Proceedings of the IEEE/CVF Conference on Computer
  Vision and Pattern Recognition. pp. 4690--4699 (2021)

\bibitem{wang2021nerf}
Wang, Z., Wu, S., Xie, W., Chen, M., Prisacariu, V.A.: Nerf--: Neural radiance
  fields without known camera parameters. arXiv preprint arXiv:2102.07064
  (2021)

\bibitem{xian2021space}
Xian, W., Huang, J.B., Kopf, J., Kim, C.: Space-time neural irradiance fields
  for free-viewpoint video. In: Proceedings of the IEEE/CVF Conference on
  Computer Vision and Pattern Recognition. pp. 9421--9431 (2021)

\bibitem{xu2019training}
Xu, Z.Q.J., Zhang, Y., Xiao, Y.: Training behavior of deep neural network in
  frequency domain. In: International Conference on Neural Information
  Processing. pp. 264--274. Springer (2019)

\bibitem{yen2021inerf}
Yen-Chen, L., Florence, P., Barron, J.T., Rodriguez, A., Isola, P., Lin, T.Y.:
  inerf: Inverting neural radiance fields for pose estimation. In: 2021
  IEEE/RSJ International Conference on Intelligent Robots and Systems (IROS).
  pp. 1323--1330. IEEE (2021)

\bibitem{yu2021plenoctrees}
Yu, A., Li, R., Tancik, M., Li, H., Ng, R., Kanazawa, A.: Plenoctrees for
  real-time rendering of neural radiance fields. In: Proceedings of the
  IEEE/CVF International Conference on Computer Vision. pp. 5752--5761 (2021)

\bibitem{yu2021pixelnerf}
Yu, A., Ye, V., Tancik, M., Kanazawa, A.: pixelnerf: Neural radiance fields
  from one or few images. In: Proceedings of the IEEE/CVF Conference on
  Computer Vision and Pattern Recognition. pp. 4578--4587 (2021)

\bibitem{yuce2021structured}
Y{\"u}ce, G., Ortiz-Jim{\'e}nez, G., Besbinar, B., Frossard, P.: A structured
  dictionary perspective on implicit neural representations. arXiv preprint
  arXiv:2112.01917  (2021)

\bibitem{zheng2021rethinking}
Zheng, J., Ramasinghe, S., Lucey, S.: Rethinking positional encoding. arXiv
  preprint arXiv:2107.02561  (2021)

\bibitem{zhu2021nice}
Zhu, Z., Peng, S., Larsson, V., Xu, W., Bao, H., Cui, Z., Oswald, M.R.,
  Pollefeys, M.: Nice-slam: Neural implicit scalable encoding for slam. arXiv
  preprint arXiv:2112.12130  (2021)

\end{thebibliography}
\end{document}